# The value of information and the emergence of communication in evolved e-puck robots

Fernando Montes-Gonzalez
*Instituto de Investigaciones en Inteligencia Artificial, Universidad Veracruzana, Mexico*

ORCID: 0000-0002-8024-3023

Correspondence: fmontes@uv.mx

## Abstract

We study when communication emerges between autonomous agents and what role an innate action selection architecture plays. Two e-puck robots, reconstructed in PyBullet from their Webots model and validated against it, must find a feeding zone that can only be detected when a robot passes over it. Each robot can emit a one-bit signal whose direction is detected by the receiver. Control is provided by a GRU recurrent network with a residual path, either operating alone or as a channel within an Embedded Action Selection Architecture (EASA). Fitness measures only the time the robots spend over the food. When the robots receive a compass pointing to the food, use of the signal disappears completely. Without a compass, in a small arena, three independent runs achieved similar performance using different strategies: a referential signal, a permanent beacon, or no use of the signal. As the arena expands, the strategy based on a referential signal generalizes best. In contrast, evolving directly in the large arena does not increase the frequency of communication, but it does separate the strategies, and the referential signal clearly outperforms solitary searching. Within EASA, innate reflexes constrain the network and reduce performance. As its priorities evolve, evolution consistently weakens those reflexes. The results suggest that communication emerges in proportion to the value of the information it transmits, and that an innate scaffolding designed for isolated agents can be an impediment in a social task.

## Keywords

emergent communication; autonomous agents; evolutionary robotics; basal ganglia; action selection; e-puck robot

## 1. Introduction

Communication allows one organism to modify the behavior of another at low cost. Understanding how a signaling system arises is difficult in animals because its origin can almost never be observed. Evolutionary robotics offers an alternative. In it, the controllers of the robots are selected based on their performance in a task, and behavior emerges from the interaction between the agent, its body, and the environment [1]. This approach has shown that simulated and physical robots can develop useful signals for foraging [2] and for group navigation [3]. This work has transformed robots into a laboratory for studying biological questions about the evolution of social behavior [4].

A central question in this line of inquiry is under what conditions communication emerges. The genetic relationship between individuals and the level at which selection operates influence the reliability of signals [2], [5]. History also plays a role. Populations that start from the same conditions can arrive at different signaling strategies due to historical contingency [6]. A more basic condition has received less attention. A signal is useful when it conveys information that the receiver cannot obtain independently. Otherwise, the signal becomes redundant, reducing the selective pressure for its maintenance.

This condition is easy to overlook when designing an experiment. When the controller receives inputs such as the direction or distance to the target, in addition to its sensors, evolution is facilitated, but communication can become unnecessary. In that case, communication signals may appear only if bonuses are added to the fitness function or rules are used to force their emission, and it becomes difficult to distinguish a signal that evolved from one that was imposed. Added to this is the gap between simulation and reality [7]. A poorly validated robot model can produce behaviors that depend on simulator artifacts rather than the task itself. In previous work, we transferred a communication protocol co-evolved in 2D directly to a 3D physical simulation without retraining, and the transfer was partial and asymmetric [8]. Here we take the opposite route. We evolve directly in a validated 3D model and ask under which conditions communication emerges at all.

A second question refers to the controller architecture. Vertebrates solve the problem of choosing what to do using the basal ganglia, a set of nuclei that arbitrates between competing motor programs [9]. Computational models of these circuits have been incorporated into robots and have produced coherent behavioral sequences in foraging tasks [10], [11]. Recently, EASA, an implementation of this model that controls a Khepera robot under realistic contact physics, was presented [12]. Such an architecture may provide a natural scaffold for evolution. Innate reflexes take care of the basics, such as obstacle avoidance, and the learned network can focus on the novel. The idea is reminiscent of layered control architectures in robots and vertebrates [13]. However, it is not obvious that an innate repertoire is helpful when the task is social. A reflex designed for an isolated agent could interfere with behaviors that depend on the partner.

In this work, we address both questions with a system built from scratch to avoid the aforementioned problems. Two e-puck robots [14], reconstructed from their official model in Webots [15] and reproduced in the PyBullet physics simulator [16], must find a food zone that they only detect by passing over it with their floor sensors. Neither robot receives information about the food's position. The only communication channel is a one-bit signal, the direction and intensity of which are perceived by the receiver. Each robot is controlled by a recurrent network with GRU units [17] and a residual connection, which acts alone or as an additional channel within the EASA basal ganglia architecture. Fitness is implicit. It only rewards the time the robots spend over the food, without rewarding transmission or approach [1].

Based on this, we manipulated two factors. The first is the need for information, which we varied by giving the robots a compass to guide them toward food and by changing the size of the arena. The second is the architecture, which ranged from the network alone to various versions of the basal ganglia, from manually set priorities to priorities that evolved along with the network. Each condition was replicated with independent runs. The signal function was evaluated with paired ablations, behavioral diagnostics, and robustness tests under conditions the robots did not encounter during evolution.

The contributions are fourfold. The first is an open model of two e-pucks with one-bit communication, designed so that other groups can replicate and extend the experiments. The second shows that the signal disappears when the information it provides becomes available through another channel. The third shows that, when communication is useful but not essential, independent runs arrive at different solutions with the same performance, and that a more difficult task does not increase the frequency of communication, but does favor the referential signal over other strategies. The fourth shows that, in this social task, the innate reflexes of the basal ganglia architecture constrain the learned network, and that the evolution of its priorities tends to diminish their influence.

The remainder of the article is organized as follows. Section 2 reviews related work. Section 3 describes the robot model, the environment, the controllers, and the evolutionary procedure. Section 4 presents the results. Section 5 discusses their implications and limitations, and Section 6 presents the conclusions.

## 2. Related work

### 2.1 Evolution of communication in robots

Early models of artificial life gave agents a dedicated signaling channel from the outset. Quinn [18] challenged this practice. A predefined channel presupposes precisely what it aims to explain, and his model showed that coordination behaviors without a dedicated channel can become communicative. Marocco and Nolfi [3] evolved simulated robots that solved a collective navigation task and developed a non-trivial signaling system. In their work, communicative and non-communicative behavior adapted to one another, such that they could not be understood separately.

Another line of research explored the conditions that favor the emergence of reliable signals. Floreano et al. [2] evolved foraging robot colonies in an environment containing both food and poison, and found that communication emerged rapidly when individuals were related or when selection operated at the colony level. When the robots' interests conflicted, selection favored the suppression of information [5], and the degree of relatedness influenced the reliability of the signals [19]. Wischmann et al. [6] further demonstrated that populations initiated under identical conditions can develop different signaling strategies due to historical contingency.

The theory of the evolution of communication adds a caveat. Sending a signal is pointless if no one responds, and responding is pointless if no one sends a signal. As a result, a population that does not communicate tends to remain that way. Scott-Phillips et al. [20] showed that this state of non-interaction is evolutionarily stable, so communication may not emerge even if it benefits both parties. In this study, we adopted the conditions that the literature identifies as favorable, a homogeneous team with team-level selection, to isolate a distinct factor: the informational value of the signal.

## 2.2 Action selection and basal ganglia in robots

A robot with multiple behaviors needs a mechanism to decide which one to execute at any given time. Layered architectures solved this problem with fixed priorities, in which higher layers suppress lower ones [21]. Redgrave et al. [9] proposed that the basal ganglia fulfill this function in vertebrates, acting as a central selector between systems competing for motor control. Gurney et al. [22] formalized this idea in a model where each action has a salience and selection occurs through disinhibition, with a selection pathway, a control pathway, and dopaminergic modulation.

This model was incorporated into a Khepera robot, where it produced coherent behavioral sequences under a wide range of sensory and motivational conditions [10]. Subsequent work with the same architecture studied the effect of reducing or increasing simulated dopamine on the robot's behavior [11]. EASA is an open reimplementation of this model that operates under realistic contact physics [12]. These studies share a fundamental idea: that robot and vertebrate control architectures organize behavior in layers with a common selection mechanism [13].

In all these cases, the basal ganglia select among hand-designed behaviors. Less attention has been given to what happens when one of the channels is a network whose behavior is acquired through evolution, and even less when the task requires coordination with another agent. This is the second question we address.

## 2.3 Simulation and design of the fitness function

Evolutionary robotics relies on simulation because evaluating thousands of controllers on physical robots is impractical. The risk is that evolution will exploit simulator flaws that do not exist in the real world. Jakobi et al. [7] showed that carefully validated simulations with adequate noise treatment reduce this gap. Simulators such as Webots offer detailed models of commercial robots [15], including the e-puck, a robot widely used in collaborative and educational robotics [14]. In this work, we reconstructed the e-puck from its Webots model within PyBullet [16] and verified each component before using it.

The design of the fitness function is the other critical decision. Nolfi and Floreano [1] distinguish between explicit functions, which reward specific aspects of behavior, and implicit functions, which only measure the outcome of the task. The former guide the search but can impose the designer's solution. In the context of basal ganglia-inspired controllers, Montes-Gonzalez et al. [23] emphasized the importance of reducing human intervention and allowing evolution to find solutions that are not entirely constructed by the designer. In a study on the emergence of communication, this risk is central because a bonus for emitting or approaching a partner makes the signal part of the objective. That is why we use an implicit fitness function that only measures the time the robots spend over the food.

## 3. Methods

### 3.1 Robot model and environment

We reconstructed the e-puck in its version 1 based on the official Webots model [15]. All measurements were copied from the model file. The body is a cylinder with a 37 mm radius and 45 mm height, with a mass of 0.15 kg, and the center of mass and inertia matrix are those of the original model. The wheels have a 20 mm radius and are 5 mm wide, and are spaced 52 mm apart by an axle. The maximum wheel angular velocity is 6.28 rad/s, resulting in a linear speed of 0.126 m/s.

Two changes were necessary to adapt the model to PyBullet [16]. Webots makes the robot slide on a small box under its body, but in PyBullet, that box drags. We replaced it with two low-friction spheres at the front and rear, half a millimeter above the floor. Additionally, the collision engine adds a margin of about 1 mm to the cylinders, causing the robot to move 5% faster than it should. Therefore, the collision radius of the wheels is 19 mm, with an effective rolling radius of 20 mm.

We validated the model against Webots' kinematics before using it. In a straight line, the robot reaches 0.1255 m/s compared to the expected 0.1256 m/s, and when turning on its axis, it reaches 4.62 rad/s compared to the expected 4.83 rad/s, a difference attributable to wheel slippage. At rest, it tilts a maximum of 1.1 degrees. The physics runs at 120 Hz, and the controller acts every 8 simulation steps, that is, every 66.7 ms. We verified that this simulation step produces virtually the same kinematics as a 240 Hz step (0.1253 m/s and 4.65 rad/s at 240 Hz) while reducing the computation time per 60 s episode from 5.75 s to 3.08 s on the machine used for all experiments. These kinematic checks can be reproduced with the validation script included in the repository (*test_step1_kinematics.py*; see Data and code availability).

The robot has eight infrared proximity sensors and three floor sensors. Its positions, orientations, response tables, and noise levels are those of Webots. The infrared sensors measure up to 7 cm and their readings range from 4,095 to 67 units. The floor sensors respond to the distance from the ground and, in our model, also to its reflectance. A normal floor produces a reading close to 856, and the darker food area produces a reading close to 171. A robot is on top of the food when the average of its three floor sensors is less than 500.

The arena is a 1 m square surrounded by 10 cm high walls. The food is a 10 cm radius circle, large enough to fit both robots. In each episode, the food and robots are randomly placed with three restrictions: each robot starts at least 30 cm from the food, the two robots are at least 15 cm apart, and neither touches a wall. An episode lasts 60 s. Table 1 summarizes the model parameters.

**Table 1** Robot parameters, simulation, and task

| Parameter | Value | Origin |
|---|---|---|
| Body radius and height | 37 mm, 45 mm | Webots Model |
| Body mass | 0.15 kg | Webots Model |
| Wheel radius, axle | 20 mm, 52 mm | Webots Model |
| Maximum wheel speed | 6.28 rad/s (0.126 m/s) | Webots Model |
| Proximity sensors | 8 infrared LEDs, 7 cm range | Webots Model |
| Floor sensors | 3 sensors, 1.6 cm range | Webots Model |
| Frequency of physics | 120 Hz | This work |
| Control period | 66.7 ms (8 steps) | This work |
| Arena | 1 m per side, 10 cm high walls | This work |
| Food zone | Circle with a radius of 10 cm | This work |
| Episode length | 60 s | This work |

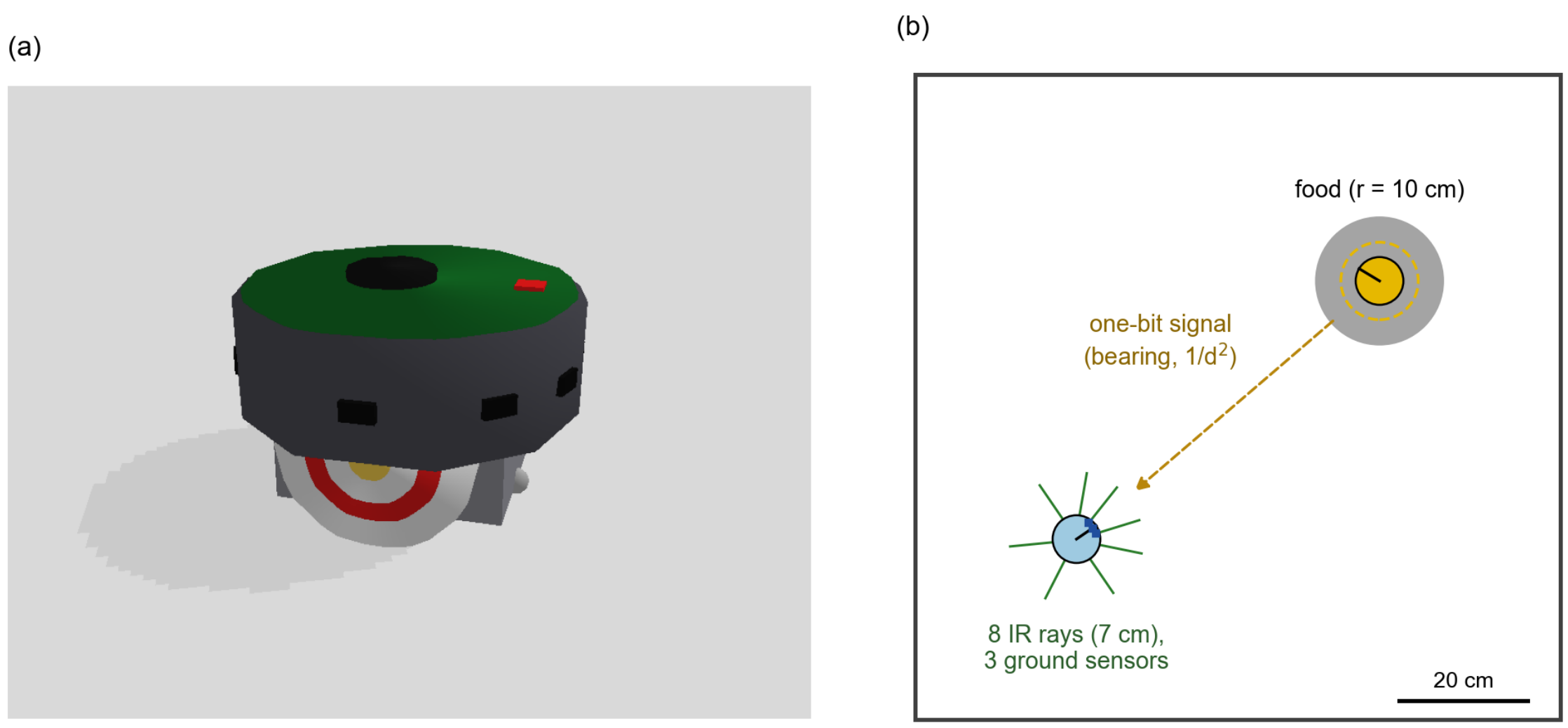

**Fig. 1** (a) Model of the e-puck reconstructed from Webots and drawn by the renderer from PyBullet. (b) Top view of the task to scale. A robot is above the food and emits the signal, and the receiver perceives its direction. The other robot displays the direction of its eight infrared sensors and the position of its three floor sensors

### 3.2 Signal

The signal was modeled after the emitter and receiver of the Webots e-puck. It is a single bit with no content. At each control step, the robot decides whether or not to emit. The receiver knows whether its partner emitted, from which direction the packet arrives in its own frame of reference, and with what strength, which decays with the square of the distance. As in Webots, a packet sent in one step arrives in the next, and the range is unlimited. This signal does not indicate the location of the food. A receiving robot only knows that its partner emitted the signal and from where. Whether the signal becomes associated with food or anything else depends entirely on evolution.

### 3.3 Control Architecture

*3.3.1 Recurrent network with residual path (Res-GRU)*

Each robot is controlled by a recurrent network with 16 GRU units [17] and a residual input-to-output connection (Figure 2). The network receives 17 inputs. The eight proximity readings are log-transformed and scaled between 0 and 1, and the three ground readings are scaled so that 0 corresponds to the floor and 1 to the food. The signal itself carries a single bit, but the receiver perceives how it arrives. Four inputs encode this percept: a reception indicator, the sine and cosine of the bearing from which the signal arrives, and its normalized strength. The sine and cosine represent the bearing without the discontinuity of an angle at ±180°. All four inputs are zero when no signal is received. The two remaining inputs are the motivations of fear and hunger.

The GRU layer follows the standard equations. The reset gate $r$ controls how much of the previous state enters the computation of a candidate state $\tilde{h}(t)$, and the update gate $z$ controls how much of the state is replaced by that candidate, so that $h(t) = (1 - z)\, h(t - 1) + z\, \tilde{h}(t)$. In Cho et al. [17] the roles of $z$ and $1 - z$ are swapped, which is equivalent to renaming the gate. The output layer sums the contribution of memory and the residual pathway, which directly connects the inputs to the outputs. This pathway allows for reactive responses that do not depend on memory.

The network has four outputs. The first two, with hyperbolic tangent activation, set the speed of each wheel as a fraction of the maximum. The third output determines whether the signal is emitted. The signal is transmitted when this value is positive. The fourth, with sigmoid activation, provides the control bid that is only used within EASA. In total, the network has 1,768 parameters.

Motivations follow the dynamics of EASA [10], [12]. Fear decreases by 0.1% at each control step, and hunger increases by 0.5%, with a maximum of 1. In the presence of food, hunger decreases by 0.5% per step. To distinguish the two robots, one starts with fear of 1.0 and hunger of 0.1, and the other with 0.9 and 0.2.

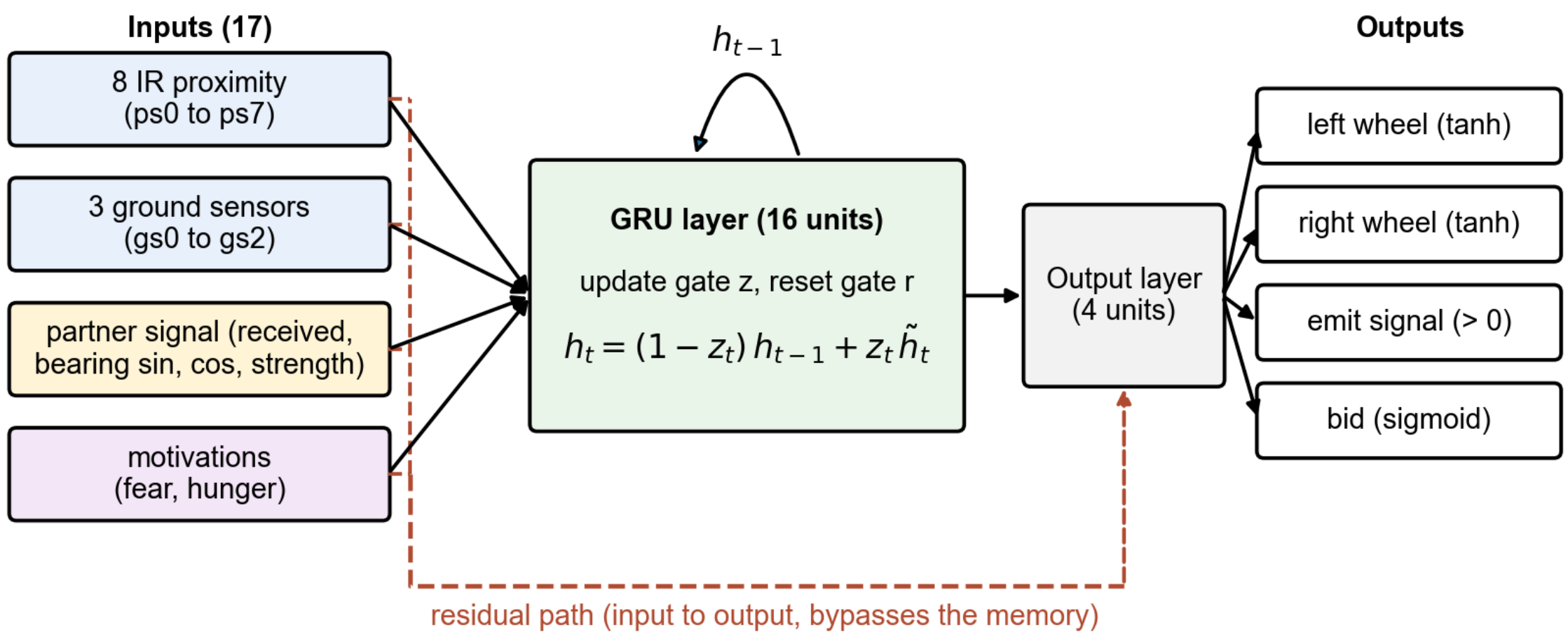

**Fig. 2** Res-GRU network architecture. The 17 inputs feed a 16-unit GRU layer with recursive memory and, via a residual path, directly to the output layer. The four outputs are the speed of each wheel, the signal transmission, and the control bid, which is only used within EASA

### *3.3.2 EASA Architecture*

EASA implements the basal ganglia model of Prescott et al. [10], based on Gurney et al. [22], with the same neurons, constants, and connections. We verified that our version reproduces the core of EASA with a maximum difference of $5 \times 10^{-15}$ over 500 steps. Figure 3 shows the system organization.

In this work, the basal ganglia choose among three channels. Two are innate. The exploration channel moves forward with a random turn that changes every one to three seconds, and the avoidance channel turns the robot away from the nearest obstacle. The third channel is the Res-GRU. The salience of each channel is a weighted sum of an *obstacle detector*, *network bid*, *fear*, and *hunger*. The basal ganglia disinhibit the channel with the highest salience, and a thalamocortical loop provides persistence for the selection. A channel passes its commands to the motor when its output from the internal globus pallidus is less than 0.4, and the commands from the selected channels are summed.

Communication is not innate in any variant. Only the network can transmit or receive the signal. We used three variants of EASA. In the first, with manually set weights, the innate channels only keep the robot safe. The second adds an innate feeding channel that stops the robot over food. In the third, the 17 selection parameters evolve along with the network, following the idea of reducing designer intervention in controllers of this type [23]. These are the 12 salience weights, the three persistence weights, and the D1 and D2 dopamine levels. This variant starts with the manually set values plus a small perturbation.

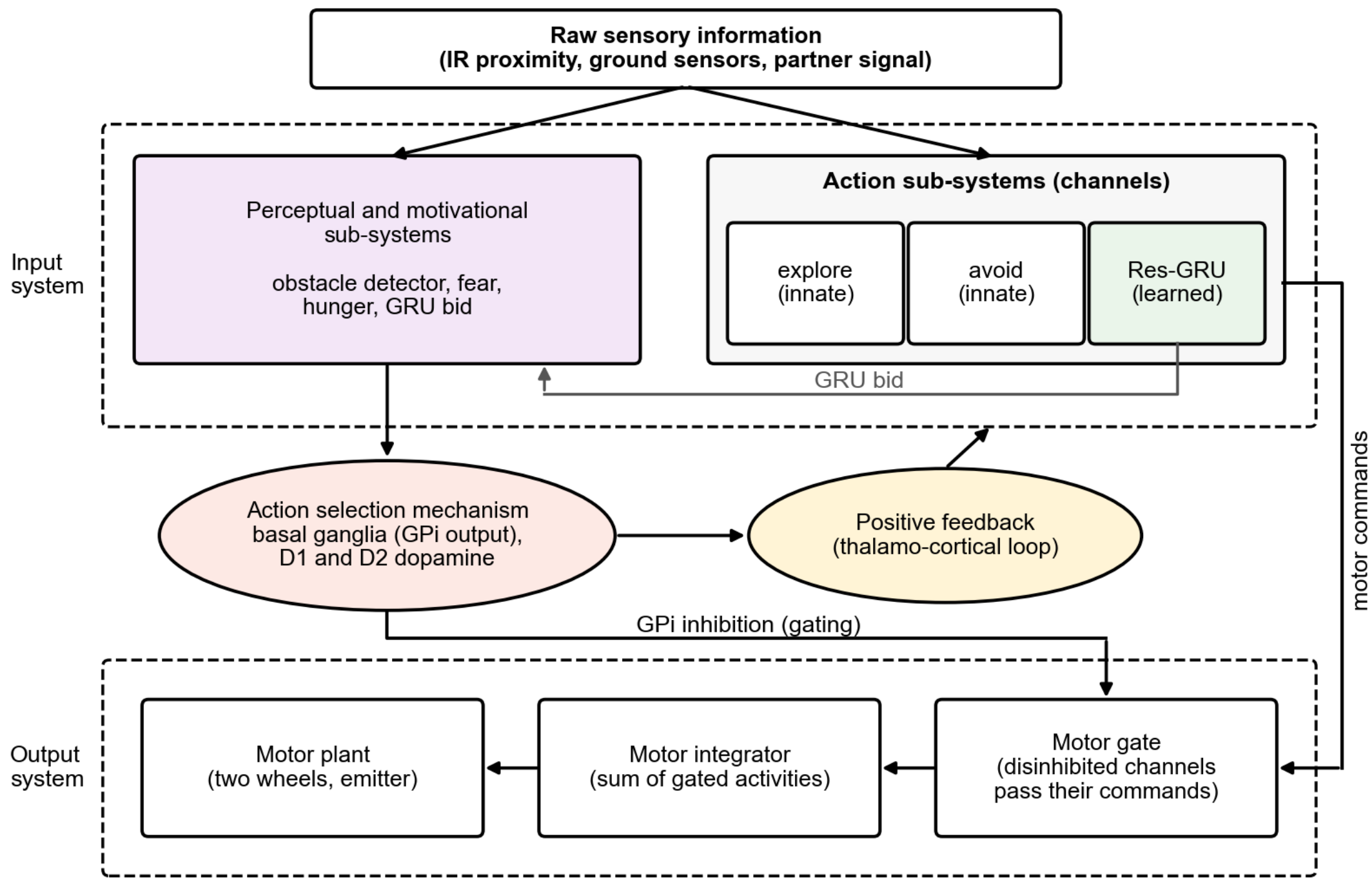

**Fig. 3** EASA architecture with the Res-GRU as a learned channel. The basal ganglia select among the innate channels (explore, avoidance) and the network, whose influence comes in as a further salient feature. The thalamocortical loop gives persistence to the selection, and the motor gate only lets through the commands of the disinhibited channels

### 3.4 Evolution and fitness function

The two robots on a team share the same genome and only differ in their initial motivations. This homogeneous team design, with selection at the team level, favors the evolution of communication [2] and leaves the informational value of the signal as the variable of interest. We use a genetic algorithm with a population of 48 individuals over 150 generations. The 10 best individuals are passed on unchanged to the next generation. The rest are obtained by uniform crossover of two parents chosen at random from the 25 best individuals, and each offspring receives Gaussian noise with a probability of 0.2 and a standard deviation of 0.12 in all its genes. The initial population is sampled from a normal distribution with a standard deviation of 0.5.

Each individual is evaluated in six episodes per generation. All individuals in a generation use the same six episodes, which change with each generation. Fitness is the fraction of the episode each robot spends over the food, averaged across both robots and the six episodes, and ranges from 0 to 1. Episodes always last a full 60 seconds. Fitness does not reward emitting or approaching the food or the partner.

Table 2 summarizes the conditions. The compass condition adds two inputs to the network: the direction and proximity to the food, information that no other condition receives. The large arena condition uses a 1.5 m square and 90 s episodes with the same food zone. All runs were performed on the same computer, a Mac mini with an M1 processor, running six processes in parallel. A run of 150 generations takes approximately six hours in the small arena condition.

**Table 2** Experimental conditions

| Condition | Controller | Innate channels | Parameters | Runs |
|---|---|---|---|---|
| GRU | Res-GRU alone | None | 1,768 | 3 |
| GRU with compass | Res-GRU plus food direction and proximity | None | 1,872 | 1 |
| EASA safe | EASA with hand-set weights | Explore, avoid | 1,768 | 1 |
| EASA eat | EASA with hand-set weights | Explore, avoid, eat | 1,768 | 1 |
| EASA Evolved | EASA with evolved priorities | Explore, avoid | 1,785 | 3 |
| GRU, large arena | Res-GRU alone | None | 1,768 | 3 |

### 3.5 Evaluation

The best individual of each generation was selected on the basis of only six episodes, so its fitness estimate is biased upward, since it may have been partly lucky. For each run, we choose a champion from the best of the last ten generations, evaluating them in 12 validation episodes that were not used during evolution. The one with the highest average fitness is the champion.

Each champion is tested in 100 new episodes with three versions of the signal. In the normal version, the signal functions as in evolution. In the disabled version, no packets arrive. In the random version, packets arrive as expected, but with a randomly chosen direction, preserving the information that the partner emitted a signal while destroying information about its location. Since all three versions use the same episodes, we compare them using the Wilcoxon signed-rank test. Comparisons between conditions with three runs use Welch's t-test on the average of each run.

The disabled version must be interpreted with caution. If a network transmits almost continuously, its peer may rely on that input as if it were constant, and removing it could disrupt the network for reasons unrelated to the information itself. Therefore, we interpret the disabled version in conjunction with the random version and signal specificity.

A behavioral assessment, conducted over 30 test episodes, measures the time spent on the food, the time spent with both robots on it, and the number of times a robot leaves the food for at least half a second. It also measures signal specificity, that is, the probability of emitting signals on the food and off it. In the EASA variants, it records which channel is active while the robot is on the food and how frequently avoidance is triggered when the partner is within 15 cm.

Finally, we tested the champions' robustness in 50 episodes under conditions they had not encountered during development. These included a 1.5 m arena with 90-second episodes, motor noise with a different gain for each wheel in every episode and additional noise at each control step, and sensor noise three times greater than that of Webots.

## 4. Results

### 4.1 Performance by condition

Figure 4 shows the evolution of fitness for all conditions in the small arena. In its three runs, GRU reached a mean population fitness close to 0.5 by the end of evolution. The EASA variants fell below this, with the hand-set priority variant having the lowest fitness. With a compass, the population reached a fitness of 0.9 in fewer than 40 generations.

Table 3 summarizes the champions' test over 100 new episodes. The three runs of GRU scored 0.588, 0.574, and 0.600, with a mean of 0.587 and a standard deviation of 0.013. The three runs of the EASA evolved condition scored 0.362, 0.356, and 0.369, with a mean of 0.362 and a standard deviation of 0.007. The difference between conditions was clear both between runs (Welch's t-test, $p = 0.0001$) and between paired episodes, where GRU outperformed EASA evolved by 0.225, equivalent to about 13.5 s more on the food per robot (Wilcoxon, $p = 5 \times 10^{-11}$). With hand-set weights, EASA eat scored 0.336 and EASA safe 0.257. GRU with compass scored 0.928.

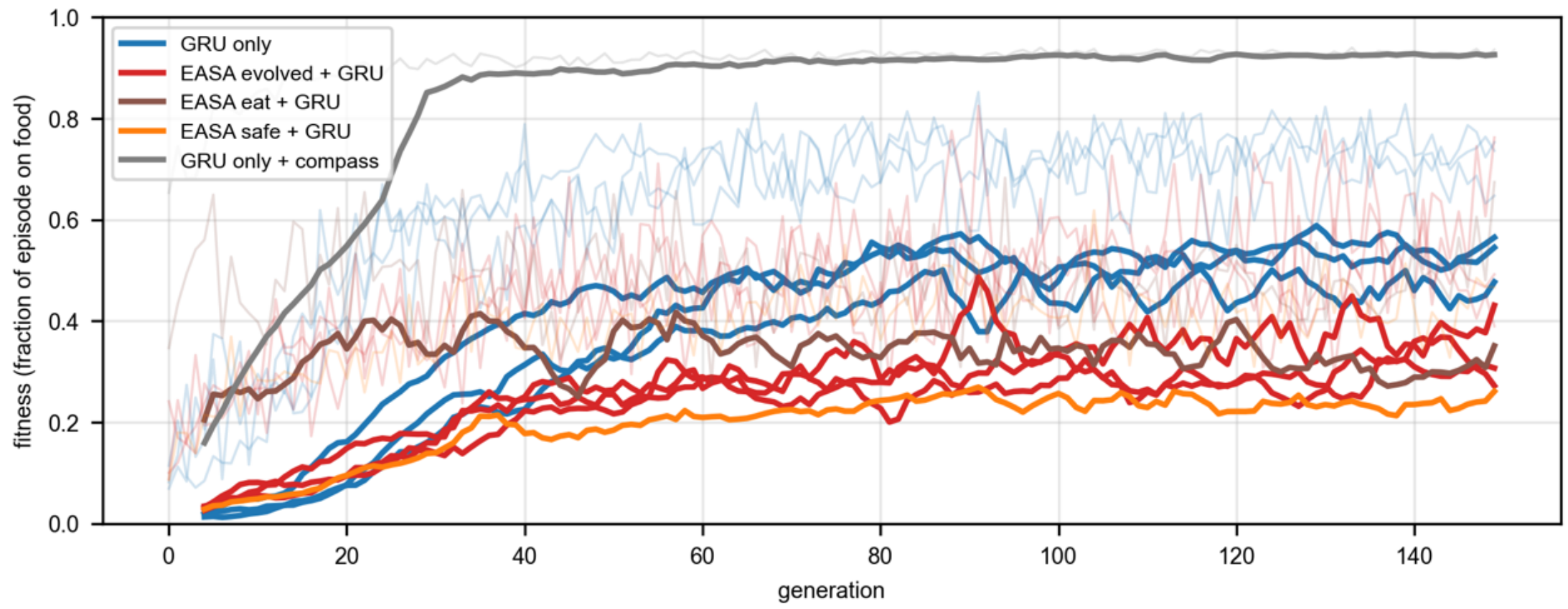

**Fig. 4** Evolution of fitness in the small arena. Thick lines show the population mean, smoothed over five generations, and thin lines show the best individual of each generation. Runs of the same condition share a color. GRU with compass reaches the highest fitness because it needs neither to search for the food nor to communicate. Among the conditions without a compass, GRU reaches the highest fitness, and one of its runs evolved the most informative signal in the study (Section 4.2)

**Table 3** Results of the champions in the small arena, 100 test episodes. Fitness under the normal, off, and random signal conditions, and signal specificity (probability of emitting on and off the food)

| Run | Normal | Off | Random | Emits on food | Emits off food | Strategy |
|---|---|---|---|---|---|---|
| GRU, run 1 | 0.588 | 0.460 | 0.353 | 96.3% | 1.3% | Referential signal |
| GRU, run 2 | 0.574 | 0.571 | 0.566 | 0.0% | 0.0% | No signal |
| GRU, run 3 | 0.600 | 0.103 | 0.203 | 82.5% | 99.8% | Beacon |
| EASA evolved, run 1 | 0.362 | 0.300 | 0.281 | 94.5% | 3.8% | Referential signal |
| EASA evolved, run 2 | 0.356 | 0.357 | 0.357 | 0.0% | 0.0% | No signal |
| EASA evolved, run 3 | 0.369 | 0.374 | 0.375 | 3.3% | 0.1% | No useful signal |
| EASA safe | 0.257 | 0.268 | 0.195 | 97.1% | 16.4% | Noisy signal |
| EASA eat | 0.336 | 0.350 | 0.347 | 0.0% | 0.6% | No signal |
| GRU with compass | 0.928 | 0.928 | 0.928 | 0.0% | 0.0% | No signal |

## 4.2 Signal Function

Paired ablations show that the signal is only used in some runs (Figure 5). In the first run of GRU, turning off the signal reduced fitness by 0.128, about 7.7 seconds over the food per robot ($p = 1 \times 10^{-7}$), and the proportion of episodes with both robots over the food dropped from 86% to 53%. Random addressing reduced it even further, to 0.353. In the third run, the drop with the signal turned off was from 0.600 to 0.103, but this figure should be read with the caveat in Section 3.5, because its transmission is almost constant. The random version, which preserves packets, lowered fitness to 0.203 and confirms that the receiver uses the direction. In the second run, all three versions of the signal yielded the same result. In the EASA evolved, only the first run used the signal, with a gain of 3.7 seconds ($p = 1 \times 10^{-4}$).

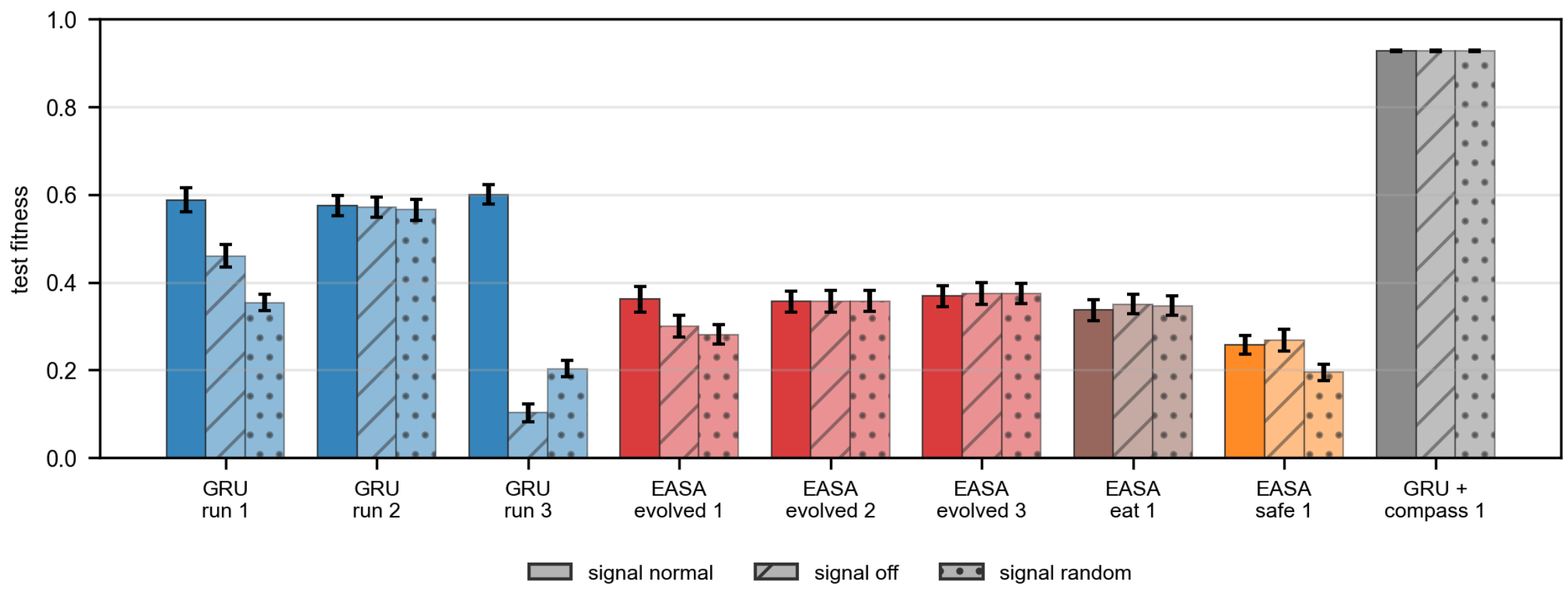

**Fig. 5** Champions' fitness in the same 100 test episodes with normal signal, signal off, and random direction. The bars indicate the mean and standard error

### 4.3 Communication strategies across runs

The runs of GRU achieved the same performance via different paths. In the first, the signal means "I am on the food". The robot emits in 96% of the control steps it spends on the food and in 1.3% of those it spends off it. In the second, there is no signal, and each robot searches on its own. In the third, the signal functions like a beacon, emitting almost always away from the food and only intermittently interrupting over it, which allows the partner to orient themselves towards the sender. The three runs do not differ from each other in fitness ($p > 0.38$ in all paired comparisons).

Behavioral analysis shows that all three strategies kept both robots on the food for a similar fraction of the episode, between 44% and 49% of the time. The network also learned to stay on the food by circling within it, and the robots almost never left it (between 0.00 and 0.73 exits per episode).

### 4.4 Compass Effect

Using a compass, the robots travel almost in a straight line towards the food (Figure 6) and both arrive in 100% of the episodes. They spend 93% of their time on it and never leave it. Population transmission fell below 1.5% by generation 65 and to zero by generation 90, and in the test, fitness was identical across the normal, off, and random signal conditions. When the information the signal could provide is available through another means, the signal disappears.

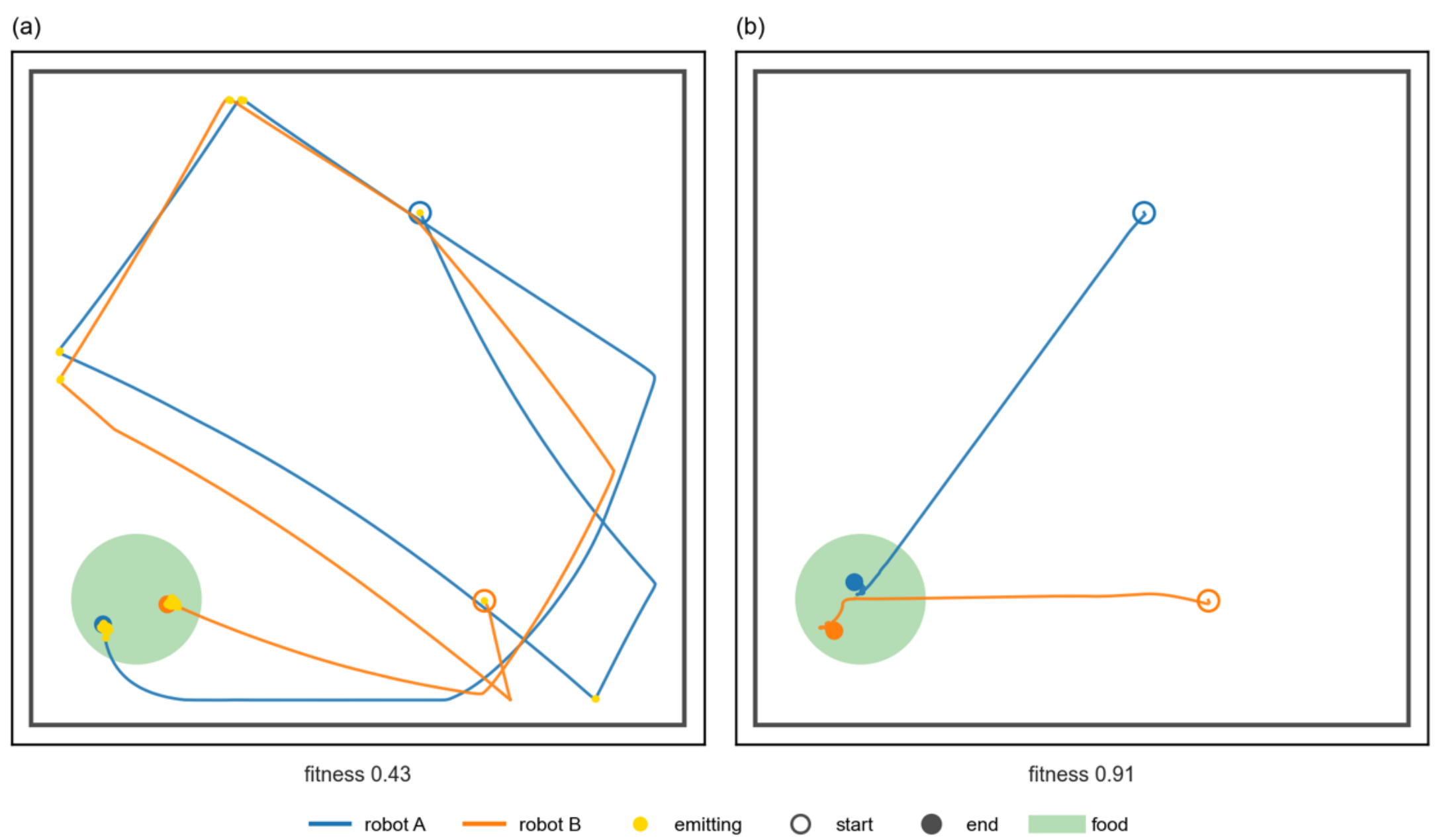

**Fig. 6** Trajectories of the two robots in the same test episode. (a) GRU, run 1, with a referential signal. (b) GRU with compass. Gold dots mark the steps in which a robot emits. Hollow circles indicate the start and filled circles the end

### 4.5 Restrictions on innate action selection

In EASA safe, with manually set weights, the robots were on the food together only 4.9% of the time and left it 1.53 times per episode. The avoidance channel was activated in 12% of the steps over the food when the partner was close and never when it was far away; that is, the reflex treated the partner as an obstacle. However, only 5 of 46 departures from the food were preceded by avoidance. The main cost appears to come from the fact that innate channels took over in about a third of the steps, almost always away from the food, and disrupted the network's search and approach strategies.

In the EASA evolved, the network was in control between 38% and 69% of the time, depending on the run, and food exits decreased to between 0.20 and 0.93 per episode. Even so, avoidance preceded a significant portion of the remaining exits. An isolated test of the architecture confirms a known property of the basal ganglia. With two channels tied under a noisy input, EASA switched channels twice in 1,000 steps, compared to 500 switches in a simple maximum selection. This persistence was not enough to compensate for the interference of reflexes in a social task.

### 4.6 Evolved priorities of the basal ganglia

Table 4 compares the evolved selection parameters in the three runs with the manually set values. Three changes appeared in all three runs. The weight of obstacle detection on avoidance decreased from 0.9 to between 0.53 and 0.79, weakening the reflex. Inhibition of the net channel by obstacles decreased from -0.3 to between -0.20 and 0.09. The hunger drive on exploration decreased from 0.6 to between -0.19 and 0.48. Other changes were not replicated. The coupling between net bidding and avoidance was positive in two runs and negative in the third, and dopamine shifted substantially in only one run. Overall, evolution reduced the weight of innate factors in all three runs.

**Table 4** EASA selection parameters, hand-set and evolved in the three runs

| **Parameter (channel ← input)** | **Hand-set** | **Run 1** | **Run 2** | **Run 3** |
|---|---|---|---|---|
| Avoid ← obstacle detector | 0.9 | 0.79 | 0.53 | 0.61 |
| Res-GRU ← obstacle detector | -0.3 | -0.20 | -0.10 | 0.09 |
| Explore ← hunger | 0.6 | 0.48 | -0.19 | 0.13 |
| Avoid ← Res-GRU bid | 0.0 | 0.53 | 0.31 | -0.81 |
| Dopamine D1, D2 | 0.20, 0.20 | 0.11, 0.28 | 0.20, 0.21 | 0.20, 0.22 |

### 4.7 Robustness and large arena

Figure 7 shows the champions' performance under conditions they did not encounter during development. Noise from the motors and sensors had minimal effects, less than 15% across all runs. The large arena significantly impacted the strategies. The referential signal run lost only 5%, the beacon run lost 23%, and the non-communicating run lost 36%. The EASA variants lost between 23% and 45%. The compass network showed no reduction in fitness.

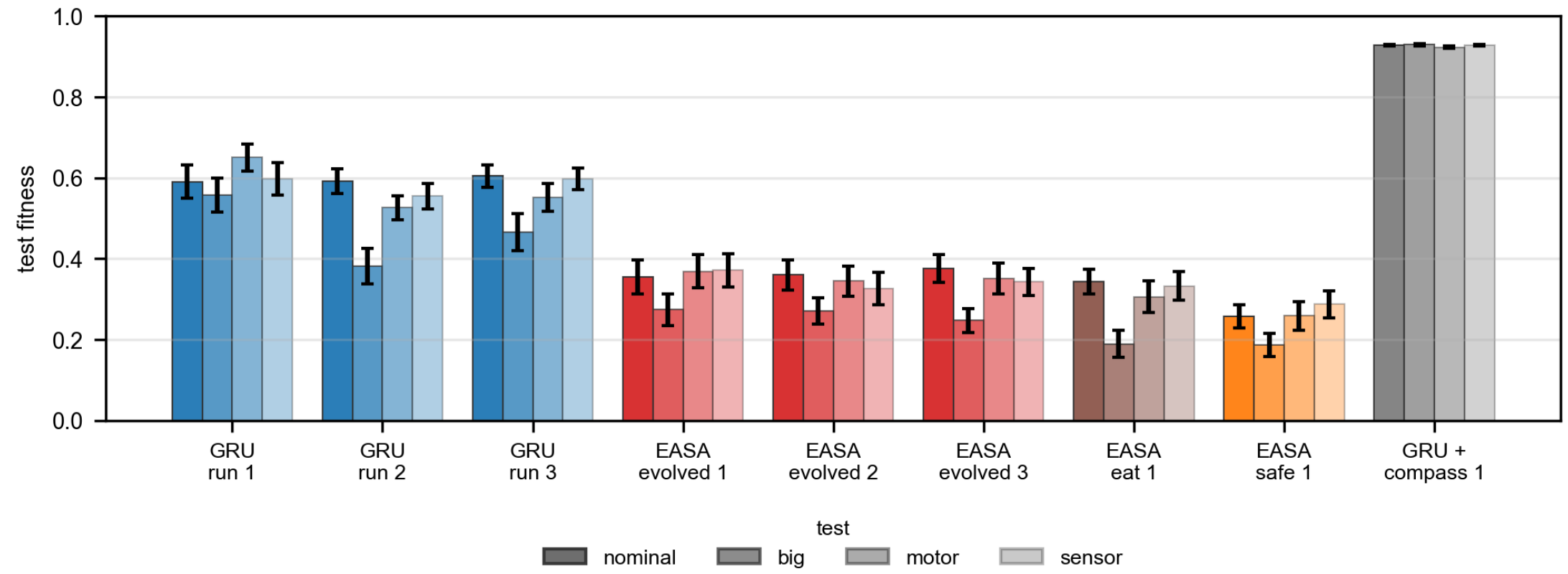

**Fig. 7** Champion fitness in the nominal condition and in three conditions not observed during evolution: a 1.5 m arena with 90 s episodes, motor noise, and tripled sensor noise. The bars indicate the mean and standard error over 50 episodes

To find out whether a more difficult task favors communication from the outset, we evolved three runs directly in the large arena (Table 5). Two communicated and one did not, the same proportion as in the small arena. The difference lay in the value of each strategy. The run with a referential signal, which emits in 99% of the steps over the food and in 2.5% of the steps off the food, scored 0.551 and outperformed the run without a signal by 0.165, about 15 seconds over the food per robot ($p = 0.0003$). The run with a beacon scored 0.229 and fell short of the one that did not communicate ($p = 0.0004$). In the small arena, all three strategies were tied, while in the large arena they were ordered from the referential to the beacon.

**Table 5** Results of the three runs evolved in the large arena, 100 test episodes of 90 s

| Run | Normal | Off | Random | Emits on food | Emits off food | Strategy |
|---|---|---|---|---|---|---|
| 1 | 0.229 | 0.082 | 0.029 | 78.3% | 97.9% | Beacon |
| 2 | 0.551 | 0.378 | 0.328 | 98.9% | 2.5% | Referential signal |
| 3 | 0.387 | 0.389 | 0.387 | 0.0% | 0.1% | No signal |

## 5. Discussion

### 5.1 Communication follows the value of information

The most direct result is that of the compass. When the robots were given the direction of the food, the signal disappeared completely, even though everything else was identical. Without a compass, however, a signal used by the other robots appeared in two out of three runs. Communication does not arise simply from having an available channel, but because it transmits something the receiver cannot know on its own. This has a practical consequence for experimental design. If the controllers receive privileged information about the target, communication can become redundant, and the only way to obtain communication is to enforce it with bonuses or rules, making it difficult to determine whether it emerged through evolution. In our earlier transfer study, the controllers received the relative bearing to the food and a distance-based term as inputs [8]. This may explain why communication there had to be supported by additional mechanisms.

However, the value of the information does not guarantee that communication will emerge. One in three runs, in both the small and large arenas, solved the task without a signal. This is consistent with the argument of Scott-Phillips et al. [20]. Sending a signal is pointless if no one responds, and responding is pointless if no one sends a signal, so a population without communication tends to remain that way. The runs that did communicate also evolved different forms, such as a referential signal or a beacon, which is reminiscent of the historical contingency described by Wischmann et al. [6].

### 5.2 When the task is more difficult, what the signal says matters

We expected a larger arena to make communication more frequent. This was not the case. What changed was the relative value of the strategies. In the small arena, the referential signal, the beacon, and the solitary search showed similar performance. In the large arena, the referential signal clearly outperformed the solitary search, and the beacon performed even worse than not communicating at all. The robustness test showed the same order with champions that had evolved in the small arena. A signal indicating the location of food retains its value when the search becomes difficult, whereas a beacon that only indicates the location of the partner can lead the receiver astray. The difficulty of the task did not create communication, but it did separate useful communication from unhelpful communication. This complements our earlier transfer study, which suggested that a communication protocol transfers well only if the ecological and navigational conditions under which it evolved are preserved [8]. Here, the referential signal evolved in the small arena kept its value in a larger one, whereas solitary search did not.

### 5.3 Innate scaffolding in a social task

The expectation that innate reflexes would facilitate evolution was not met. All three EASA variants performed worse than GRU and were also less robust in the large arena. The diagnosis points to two mechanisms. The avoidance reflex treats the partner as an obstacle precisely when the two should be together, and innate channels disrupt the strategies the network develops to search for and approach the partner. In this task, a repertoire designed for an isolated agent proved ill-suited when success depended on another agent.

Evolution's response was consistent. When allowed to adjust priorities, it weakened the avoidance reflex, protected the network from obstacle inhibition, and reduced the drive to explore, always in the direction of yielding control to what has been learned. This supports the idea of reducing designer intervention in basal ganglia-inspired controllers [23] and qualifies the idea of layered architectures with fixed priorities [13], [21]. The persistence provided by the basal ganglia was confirmed in the isolated test, but it was not enough to compensate for the interference. One possibility we did not explore is that the basal ganglia select among several learned proposals rather than among fixed reflexes, which would be closer to the vertebrate organization, in which the cortex proposes and the basal ganglia select [9].

### 5.4 Limitations

Each main condition has three runs. This is sufficient to show large and consistent differences, but not to accurately estimate the frequency of each communication strategy. Conclusions about behavior are based on the champion of each run, not the entire population. All results are from simulations, although with a model validated against Webots. Simulation physics are not identical across computers, so all runs were performed on the same machine. Finally, we used a single task, a single network form, and a single evolutionary algorithm, and the EASA variant explores only one way of combining innate reflexes with a learned network.

## 6. Conclusions

We built an open, validated system of two e-puck robots to study the emergence of a one-bit signal. With it, we showed three things. First, communication disappears when the information it provides is available through another channel. When it is not, communication appears in most runs, though not in all of them. Second, a more difficult task does not make communication more frequent, but it does distinguish the referential signal, which retains its value, from less informative strategies. Third, in a social task, the innate reflexes of a basal ganglia architecture constrain the learned network, and evolution consistently responds by shifting control to the learned network. A natural next step is to test an architecture in which the basal ganglia select among learned proposals, and to verify these results with physical robots.

## Acknowledgments

The author is grateful for the support of Universidad Veracruzana. Fernando Montes-Gonzalez is a member of the National System of Researchers (SNII), Mexico.

## Declarations

### Funding

No funding was received for this study.

### Conflict of interest

The author declares no conflicts of interest.

### Ethical approval

Not applicable. This study does not involve human participants or animals.

### Author contribution

F. Montes-Gonzalez conceived the study, developed the software, conducted the experiments and analysis, and wrote the manuscript.

### Data and code availability

The source code, evolution and evaluation scripts, and the datasets required to reproduce the results are publicly available on GitHub at https://github.com/ferdiex/easagru and archived on Zenodo at https://doi.org/10.5281/zenodo.23051976.